\documentclass[letterpaper]{article} 
\usepackage{aaai25}  
\usepackage{times}  
\usepackage{helvet}  
\usepackage{courier}  
\usepackage[hyphens]{url}  
\usepackage{graphicx} 
\usepackage{natbib}  
\usepackage{caption} 
\usepackage{algorithm}
\usepackage[utf8]{inputenc} 
\usepackage[T1]{fontenc}    
\usepackage{url}            
\usepackage{booktabs}       
\usepackage{amsfonts}       
\usepackage{nicefrac}       
\usepackage{microtype}      
\usepackage{xcolor}         
\usepackage{tcolorbox} 
\usepackage{natbib}
\usepackage{enumitem}
\usepackage{xcolor}
\usepackage{tikz}
\usepackage{pgfplots}
\usepackage[multiple]{footmisc}
\usepackage{multirow}
\usepackage{bm}
\usepackage{booktabs}
\usepackage{array}
\usepackage{subcaption}
\usepackage{graphicx}
\usepackage{tabularx}
\usepackage{lipsum}
\usepackage{algpseudocode}
\usepackage{soul}
\usepackage[flushleft]{threeparttable}
\usepackage{tablefootnote}
\usetikzlibrary{pgfplots.groupplots}
\usepackage[english]{babel}
\usepackage{amsthm}
\usepackage{amsmath}

\usepackage{amssymb}
\usepackage{pifont}
\usepackage{pdfrender}

\pgfplotsset{
    name nodes near coords/.style={
        every node near coord/.append style={
            name=#1-\coordindex,
            alias=#1-last,
        },
    },
    name nodes near coords/.default=coordnode
}

\pgfplotsset{compat=1.5.1}
\def\addlegendimage{\csname pgfplots@addlegendimage\endcsname}
\usetikzlibrary{patterns}
\usetikzlibrary{pgfplots.groupplots}
\usetikzlibrary{
  pgfplots.colorbrewer,
}

\definecolor{aa}{rgb}{0.2,0.7,0.310}
\definecolor{cc}{rgb}{1.0,0.49,0.0}
\definecolor{bb}{rgb}{0.514,0.325,0.831}

\usepackage{newfloat}
\usepackage{listings}
\DeclareCaptionStyle{ruled}{labelfont=normalfont,labelsep=colon,strut=off} 
\floatstyle{ruled}
\newfloat{listing}{tb}{lst}{}
\floatname{listing}{Listing}
\title{Disentangling, Amplifying, and Debiasing: \\ Learning Disentangled Representations for Fair Graph Neural Networks}

\author {
    Yeon-Chang Lee\textsuperscript{\rm 1},
    Hojung Shin\textsuperscript{\rm 2},
    Sang-Wook Kim\textsuperscript{\rm 2}\thanks{Corresponding author.} 
}
\affiliations {
    \textsuperscript{\rm 1}Ulsan National Institute of Science and Technology (UNIST)\\
    \textsuperscript{\rm 2}Hanyang University\\
    yeonchang@unist.ac.kr, \{hojungshin, wook\}@hanyang.ac.kr
}

\newcommand{\ie}{{\textit i.e.}}
\newcommand{\eg}{{\textit e.g.}}
\newcommand{\ours}{\textbf{\texttt{DAB}-\texttt{GNN}}}
\newcommand{\fagnn}{FGNN}

\newcommand{\abdisen}{\textit{AbDisen}}
\newcommand{\abemb}{\textit{AbEmb}}
\newcommand{\sbdisen}{\textit{SbDisen}}
\newcommand{\sbemb}{\textit{SbEmb}}
\newcommand{\pbdisen}{\textit{PbDisen}}
\newcommand{\pbemb}{\textit{PbEmb}}

\begin{document}

\maketitle

\begin{abstract}
Graph Neural Networks (GNNs) have become essential tools for graph representation learning in various domains, such as social media and healthcare. 
However, they often suffer from fairness issues due to inherent biases in node attributes and graph structure, leading to unfair predictions.
To address these challenges, we propose a novel GNN framework, \ours, that \ul{D}isentangles, \ul{A}mplifies, and de\ul{B}iases attribute, structure, and potential biases in the \ul{GNN} mechanism. 
\ours\ employs a disentanglement and amplification module that isolates and amplifies each type of bias through specialized disentanglers, followed by a debiasing module that minimizes the distance between subgroup distributions.
Extensive experiments on five datasets demonstrate that \ours\ significantly outperforms ten state-of-the-art competitors in terms of achieving an optimal balance between accuracy and fairness.
\end{abstract}

\begin{links}
\link{Code}{https://github.com/Bigdasgit/DAB-GNN}
\end{links}

\section{Introduction}\label{sec:introduction}

\textbf{Background}.
In the real world, data from various domains such as social media, healthcare, and finance can be represented as graph data \cite{RozemberczkiHGG22, TangL09, YooLSK23, KimHKSK24}.
In these graphs, entities (\eg, users) are depicted as nodes, and pairwise relationships between these entities (\eg, friendship) are depicted as edges. This structural representation enables the effective analysis of complex relationships within the graph data.

Recent advancements in graph representation learning have utilized \textit{Graph Neural Networks} (GNNs) to map nodes into low-dimensional embedding space \cite{KipfW17, KimLK23, SharmaLNSSKK24}.
Using a message-passing framework,
GNNs iteratively aggregate information from a node and its neighbors 
to produce the final embedding of the node that effectively captures both node attributes and the graph's structure \cite{HamiltonYL17, KipfW17}.
Consequently, GNNs have demonstrated superior performance in a variety of tasks like node classification and link prediction \cite{SunLWSLW24, ChamberlainSRFM23, KimLK24}.

\vspace{1mm}
\noindent\textbf{Motivation}.
However, predictions based on node embeddings learned by GNNs can be \textit{unfair} due to the biases inherent in the input graph data, such as node attributes and graph structure, as well as the biases introduced by the message-passing mechanism of GNNs \cite{DaiW21, WangZDCLD22, LingJLJZ23, LiWXFWLS24, JiangHFLZMH24, NeoLJKK24}.
Specifically, node attributes may exhibit different distributions across subgroups (\eg, male and female) defined by \textit{sensitive attributes} (\eg, gender), referred to as \textit{attribute bias} \cite{DongLJL22}.
For instance, in job application data, the average income of employees may differ across genders.
Additionally, the graph structure itself can be biased, as nodes with similar sensitive attributes tend to form connections, known as \textit{structure bias} \cite{DongLJL22}. 
For example, on social network platforms, users predominantly form friendships within same sensitive attributes \cite{RahmanS0019, DaiW21}.

Moreover, the message-passing mechanism in GNNs can introduce what we call \textit{potential bias} by combining
existing attribute and structure biases.
When node attributes and graph structure interact, new biases may emerge that were not evident in either aspect alone.
As one instance of potential bias, \citet{WangZDCLD22} observed that biases related to sensitive attributes can unintentionally spread to non-sensitive attributes during the GNN process.
This type of bias is distinct as it arises from the interplay between node attributes and graph structure.\footnote{It might be an amplified form of attribute or structure bias, or something entirely new. However, given the uncertainty of its nature, we refer to it as potential bias in this paper.}
As a result, a GNN method may inadvertently encode these biases in the final embeddings, leading to unfair predictions correlated with sensitive attributes. 

\vspace{1mm}
\noindent\textbf{Challenges}.
To mitigate these issues, various fairness-aware GNN (\fagnn) methods, such as FairGNN~\cite{DaiW21}, EDITS~\cite{DongLJL22}, and FairVGNN~\cite{WangZDCLD22}, have been proposed~\cite{DongMWCL23}, which will  be discussed in detail in Section~\ref{sec:pre}.
They are generally classified into three categories: pre-processing, in-processing, and post-processing \cite{ChenRPTWYKDA24}.
The pre-processing approach aims to eliminate biases \textit{before} model training, while 
the in-processing approach modifies the objective function or model architecture aiming to learn bias-free node embeddings \textit{during} training. 
The post-processing approach aims to adjust the final embeddings or predictions \textit{after} training.

However, existing methods often overlook a critical aspect in removing sensitive information from the final node embeddings:
``\textit{Not All Biases Are the Same.}''
Attribute bias, structure bias, and potential bias each causes sensitive attributes to affect the model: (i) attribute bias affects how node attributes are distributed across subgroups; (ii) structure bias stems from connections between nodes with similar sensitive attributes; (iii) potential bias arises when the interplay between node attributes and graph structure 
makes neutral attributes strongly correlated with sensitive attributes.
Recently, some methods have attempted to identify and address specific biases (\eg, attribute and structure biases for EDITS~\cite{DongLJL22}).
However, all \fagnn\ methods, including EDITS, still try to tackle all biases \textit{at once} by using a \textit{single, entangled embedding} for each node. 
This \textit{one-size-fits-all} strategy may fail to address the unique nature of each bias, leading to inadequate debiasing and persistent unfairness.
Consequently, effectively disentangling these biases within node embeddings remains a significant challenge.

\vspace{1mm}
\noindent\textbf{Our Work}.
To address the challenge, we propose a novel method named \textbf{\ours}, which \ul{D}isentangles, \ul{A}mplifies, and de\ul{B}iases the attribute, structure, and potential biases through a \ul{GNN} framework.
\ours\ operates with two key modules: disentanglement and amplification, and debiasing.

The \textbf{disentanglement and amplification} module uses \textit{three disentanglers}, each dedicated to isolating a specific type of bias--attribute, structure, or potential--from the input graph, and encoding them into three disentangled embeddings for each node.
In this module, we \textit{deliberately amplify} these biases by leveraging the message-passing mechanism of GNN. 
This amplification preserves the distinct property of each bias, significantly enhancing the effectiveness of the subsequent debiasing process.
The \textbf{debiasing} module employs two regularizers: the \textit{bias contrast optimizer} (BCO) and the \textit{fairness harmonizer} (FH). 
The BCO ensures that different bias embeddings remain clearly distinct, while the FH reduces the impact of sensitive attributes by aligning subgroup distributions within each bias-specific embedding.

Once the disentangled embeddings have been created, they are concatenated and used to train model parameters for various downstream tasks like node classification and link prediction.
Extensive experiments demonstrate that \ours\ successfully captures different biases from the input graph, and its strategies for disentangling, amplifying, and debiasing these biases are highly effective in mitigating unfairness.

\vspace{1mm}
\noindent\textbf{Contributions}.
Our contributions are as follows:

\begin{itemize}[leftmargin=*]
    \item \textbf{Observation:}
    We identify and address the critical limitations of learning entangled embeddings in GNNs, highlighting their impact on fairness.
    \item \textbf{Novel Framework:}
    We introduce \ours, a framework that learns disentangled embeddings for fair GNNs.
       \begin{itemize}[leftmargin=*]
       \item We design a three-disentangler architecture that effectively isolates and amplifies attribute, structure, and potential biases in the embedding space.
       \item We devise a multi-objective loss function that further separates these disentangled biases while minimizing the influence of sensitive attributes in the embeddings.
       \end{itemize}
    \item \textbf{Experimental Validation:}
    We validate \ours\ by comparing it with ten state-of-the-art competitors across five real-world datasets, achieving a superior trade-off between accuracy and fairness metrics.
\end{itemize}

\section{Preliminaries}\label{sec:pre}

\subsection{Related Work}
Fairness in graph mining, particularly in the context of GNNs, is a crucial area of research. 
Commonly studied notions in \fagnn\ include group fairness and individual fairness \cite{DuYZH21}.
\textbf{Group fairness} ensures that an algorithm does not produce biased outcomes against minority groups defined by sensitive attributes
On the other hand, \textbf{individual fairness} ensures that an algorithm gives similar outcomes to similar nodes.
We focus on group fairness, the most widely studied concept in \fagnn\ research, and review recent studies proposed to ensure group fairness in GNNs. 

FairGNN~\cite{DaiW21} uses an adversarial network to remove sensitive information from embeddings and designs a sensitive attribute estimator when such information is limited.
EDITS~\cite{DongLJL22} modifies graph data to reduce the Wasserstein distance between subgroups before training, while FairVGNN~\cite{WangZDCLD22} minimizes sensitive attribute leakage through adversarial networks.
PFR-AX~\cite{MerchantC23} uses Pairwise Fair Representation (PFR) technique~\cite{LahotiGW19} to decrease subgroup separability, and
PostProcess~\cite{MerchantC23} adjusts outcomes to close the prediction gaps for minorities.
BIND~\cite{DongW0LL23} introduces Probabilistic Distribution Disparity (PDD) to measure and remove bias-contributing nodes.

In contrast, methods like NIFTY, CAF, and GEAR leverage counterfactuals to ensure group fairness.
NIFTY~\cite{AgarwalLZ21} enhances fairness and stability by utilizing edge drops, attribute noise, and counterfactuals that flip sensitive attributes. 
CAF~\cite{Guo0X0W23} generates counterfactuals by finding similar nodes from other subgroups to minimize discrepancies. 
GEAR~\cite{MaGWYZL22} uses GraphVAE~\cite{KipfW16a} to generate counterfactuals when sensitive attributes are altered, reducing the gap between the original and counterfactual embeddings.

\subsection{Fairness Analysis on Real-World Datasets}

We begin by reviewing the fairness metrics commonly used in \fagnn\ research, followed by an analysis of real-world graph datasets based on these metrics.

\vspace{1mm}
\noindent\textbf{Embedding-Level Fairness Metrics}.
A key measure of bias in GNNs is the distribution difference of node attributes or learned embeddings between subgroups \cite{DongLJL22}. 
A greater distribution difference can lead to more-biased outcomes by making it easier for GNNs to infer a node's sensitive attribute \cite{BuylB20, ChenRPTWYKDA24}. 
To assess this, we use two metrics related to distribution difference, which are proposed by~\citet{DongLJL22}:

\begin{itemize}[leftmargin=*]
    \item \textbf{Attribute Bias (AttrBias)} measures the distribution difference in the attribute matrix between subgroups.

    \item \textbf{Structure Bias (StruBias)} measures the distribution difference of node embeddings between subgroups after applying a GNN method.

\end{itemize}

\begin{table}[t]
\centering

\renewcommand{\arraystretch}{1.1}
\resizebox{.38\textwidth}{!}{
\begin{tabular}{c|cccc}
\toprule
\multirow{2}{*}{\textbf{Datasets}} & \multirow{2}{*}{\textbf{Nodes}} & \multirow{2}{*}{\textbf{Edges}} & \multirow{2}{*}{\begin{tabular}[c]{@{}c@{}}\textbf{Intra-Group} \\ \textbf{Edges}\end{tabular}} & \multirow{2}{*}{\begin{tabular}[c]{@{}c@{}}\textbf{Inter-Group } \\ \textbf{Edges}\end{tabular}} \\
& & & & \\
\midrule
\textbf{NBA}         & 403    & 10,822    & 7,887             & 2,935             \\
\textbf{Recidivism}  & 18,876 & 321,308   & 172,259           & 149,049           \\
\textbf{Credit}      & 30,000 & 1,436,858 & 1,379,322         & 57,536            \\
\textbf{Pokec\_n}    & 66,569 & 550,331   & 526,038           & 24,293            \\
\textbf{Pokec\_z}    & 67,797 & 651,856   & 621,337           & 30,519            \\
\bottomrule
\end{tabular}
}
\caption{Dataset statistics.}
\label{tab:dataset}
\end{table}

\noindent\textbf{Neighborhood-Level Fairness Metrics}.
In GNNs, the aggregation of neighborhood information heavily influences node embeddings \cite{WuPCLZY21}.
When intra-subgroup connections dominate and inter-subgroup connections are sparse, GNNs tend to learn embeddings that are similar within subgroups and distinct across subgroups \cite{ChenRPTWYKDA24, WangZDCLD22, DaiW21}.
We measure this effect by using two neighborhood-level metrics:

\begin{itemize}[leftmargin=*]
    \item \textbf{Homophily Ratio (HomoRatio)} measures the proportion of intra-subgroup edges to all edges in a graph dataset.

    \item \textbf{Neighborhood Fairness (NbhdFair)} captures the average entropy of each node's neighbors in a graph dataset.

\end{itemize}
Detailed equations for calculating each of these metrics can be found in the online appendix at \url{https://github.com/Bigdasgit/DAB-GNN}.

Using the metrics discussed above, we analyzed the inherent biases in various real-world graph datasets \cite{DaiW21, takac2012data, AgarwalLZ21}, including
NBA, Recidivism, Credit, Pokec\_z, and Pokec\_n, detailed in Table~\ref{tab:dataset}.

\begin{itemize}[leftmargin=*]
    \item \textbf{NBA:} This dataset includes NBA player demographics and Twitter friendships. The sensitive attribute is nationality (American or not), and the task is to predict whether a player earns above the median salary.
    \item \textbf{Recidivism:} This dataset contains defendants released on bail, with relationships based on crime records and demographics. The sensitive attribute is race, and the task is to predict bail eligibility.
    \item \textbf{Credit:} This dataset includes credit card users, with relationships based on payment similarity. The sensitive attribute is age, and the task is to predict credit card default.
    \item \textbf{Pokec\_n and Pokec\_z:} These datasets are from the Slovak social network Pokec, divided by region, with friendships forming the relationships. The sensitive attribute is region, and the task is to predict the user's working field.
\end{itemize}

Table~\ref{tab:fairness_analysis} presents the results of the fairness metrics. Higher values for attribute bias and structure bias, along with a homophily ratio closer to 1 and neighborhood fairness closer to 0, indicate a higher degree of inherent bias in each dataset.
The analysis reveals that biases vary across datasets, with each exhibiting different dominant biases.
For example, the NBA dataset shows significant attribute and structure biases, high homophily, and moderate neighborhood fairness,  
while the Pokec datasets show low attribute and structure biases but high homophily and low neighborhood fairness
These findings support our claim in Section~\ref{sec:introduction}, emphasizing the importance of addressing each bias according to its unique characteristics, as biases manifest differently and do not always follow consistent patterns in different datasets.

\begin{table}[t]
\centering

\renewcommand{\arraystretch}{1.15}
\resizebox{.43\textwidth}{!}{
\begin{tabular}{c|cccc}
\toprule
\textbf{Datasets} & \textbf{AttrBias ($\uparrow$)} & \textbf{StruBias ($\uparrow$)} & \textbf{HomoRatio ($\uparrow$)} & \textbf{NbhdFair ($\downarrow$)} \\
\midrule
\textbf{NBA}          & 4.148  & 5.898  & 0.729  & 0.499 \\
\textbf{Recidivism}   & 0.953  & 1.098  & 0.536  & 0.657 \\
\textbf{Credit}       & 2.463  & 4.451  & 0.960  & 0.158 \\
\textbf{Pokec\_n}     & 0.142  & 0.248  & 0.956  & 0.114 \\
\textbf{Pokec\_z}     & 0.009  & 0.179  & 0.953  & 0.132 \\
\bottomrule
\end{tabular}
}
\caption{Fairness analysis for graph datasets. ($\uparrow$) and ($\downarrow$) indicate that higher and lower values correspond to greater inherent bias for the corresponding metric, respectively.}
\label{tab:fairness_analysis}
\end{table}

\begin{figure*}[t]
\centering
\includegraphics[width=0.96\linewidth]{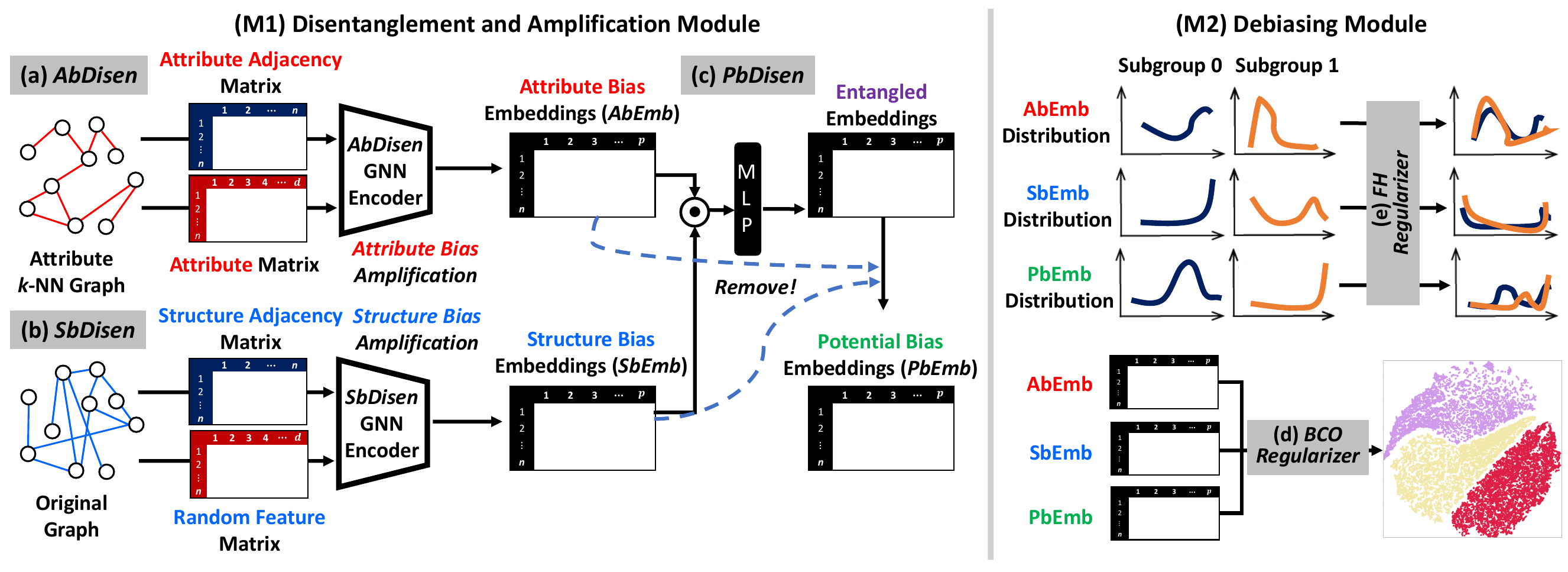}
\caption{Overview of \ours, which consists of (M1) disentanglement and amplification module, and (M2) debiasing module.}

\label{fig:overview}
\vspace{-0.45cm}
\end{figure*}

\section{The Proposed Framework: \ours}
\label{sec:approach}

\subsection{Problem Definition}

Given an attributed graph $\mathcal{G} = (\mathcal{V}, \mathcal{E}, \mathbf{X})$, where $\mathcal{V}$ and $\mathcal{E}$ denote the sets of $n$ nodes and $m$ edges, respectively, and $\mathbf{X} \in \mathbb{R}^{n\times d}$ represents the node attribute matrix with $d$ attributes, GNNs aim to learn $p$-dimensional node embeddings $\mathbf{H} \in \mathbb{R}^{n\times p}$ that capture the graph's structure and attribute information. 
However, the presence of a sensitive attribute (\eg, gender) in the nodes can introduce biases that lead to unfair predictions. 
This attribute, which can be represented as $\mathbf{S} \in \{0,1\}^{n}$ if it is binary, may cause the embeddings to inadvertently encode discriminatory patterns against certain subgroups. 
Thus, the goal of \fagnn\ methods is to learn node embeddings that are both fair and accurate, \ie, removing biases related to the sensitive attribute while maintaining the accuracy of downstream tasks.

\subsection{Overview}

Figure~\ref{fig:overview} provides an overview of \textbf{\ours}, and Table~\ref{tab:notations} lists the notations used in this paper.
\ours\ consists of two key modules: (M1) Disentanglement and Amplification Module, and (M2) Debiasing Module. 
In \textbf{(M1)}, \ours\ disentangles node embeddings into three components: attribute bias, structure bias, and potential bias. 
Each component is handled by a specialized disentangler that identifies and amplifies the corresponding bias.
These disentangled embeddings are then concatenated into a comprehensive representation, which is used for training in various downstream tasks like node classification or link prediction.
In \textbf{(M2)}, \ours\ refines the disentangled embeddings to ensure they are distinct and fair. 
This is achieved through two key regularizers: the bias contrast optimizer (BCO), which enforces clear separation between different bias embeddings, and the fairness harmonizer (FH), which reduces the impact of sensitive attributes by minimizing the distance between subgroup distributions.

\subsection{Key Modules}

\vspace{1mm}
\noindent\textbf{(M1) Disentanglement and Amplification Module}.
The goal of this module is to disentangle the node embeddings into three distinct components, each addressing a specific type of bias: attribute bias, structure bias, and potential bias. 
As discussed in Section~\ref{sec:introduction}, biases in graph data can originate from various sources, including the node attributes, the graph structure, or the interactions between these two elements.
By separating these biases, \ours\ can independently address each one, making it more effective in identifying and mitigating their effects.

\begin{table}[t]
\centering

\resizebox{.48\textwidth}{!}{
\renewcommand{\arraystretch}{1.1}
\begin{tabular}{p{0.35\linewidth}|p{0.9\linewidth}}
\toprule
\hfil\textbf{Notation} & \textbf{Description} \\ \midrule
\hfil$\mathcal{G} = (\mathcal{V}, \mathcal{E}, \mathbf{X})$ & Attributed graph with node set $\mathcal{V}$, edge set $\mathcal{E}$, and node feature matrix $\mathbf{X}$ \\ 
\hfil$\mathbf{S} \in \{0,1\}^{n}$ & Sensitive attribute that may introduce bias \\ 
\hfil$\mathbf{A}_\text{attr},\mathbf{A}_\text{stru}$ & Adjacency matrices of the $k$-NN graph based on attribute similarity and the input graph. \\ \midrule
\hfil$\mathbf{H}_\text{attr}$, $\mathbf{H}_\text{stru}$, $\mathbf{H}_\text{pot}$   & Disentangled node embeddings for attribute, structure, and potential biases  \\  
\hfil$\mathcal{L}_\text{primary}$, $\mathcal{L}_\text{bco}$, $\mathcal{L}_\text{fh}$ & Losses for the primary downstream task, the bias contrast optimizer, and the fairness harmonizer  \\ 
\hfil$\mathcal{L}_\text{total}$ & Final loss function combining $\mathcal{L}_\text{primary}$, $\mathcal{L}_\text{bco}$, and $\mathcal{L}_\text{fh}$ \\ 
\hfil$\alpha, \beta$ & Weights for $\mathcal{L}_\text{fh}$ and $\mathcal{L}_\text{bco}$ \\
\bottomrule
\end{tabular}
}
\caption{Notations used in this paper.}
\label{tab:notations}
\end{table}

\begin{itemize}
    \item \textbf{Attribute Bias Disentangler (\abdisen)}: 
    This component leverages a specialized GNN to effectively capture the attribute bias. 
    The process begins with the attribute matrix $\mathbf{X}_\text{attr} \in \mathbb{R}^{n \times d}$, which represents the node attributes of the input graph. 
    Using $\mathbf{X}_\text{attr}$, we construct a $k$-Nearest Neighbors ($k$-NN) graph based on the Euclidean distance between node attributes.
    The distance between two nodes $i$ and $j$ is quantified as $\texttt{D}_{e}(i, j) = \sqrt{\sum_{ r=1}^{d} (x_{ir} - x_{jr})^2}$,
    where ${x}_{i}$ and ${x}_{j}$ represent the attribute vectors for nodes $i$ and $j$ in $\mathbf{X}_\text{attr}$, respectively. 
    Based on $\texttt{D}_{e}(i, \cdot)$, the $k$ nearest neighbors for each node $i$ are identified. 
    The $k$-NN graph is then represented by the adjacency matrix $\mathbf{A}_\text{attr} \in \mathbb{R}^{n \times n}$:
    
    \small
    \begin{equation}
    \mathbf{A}_\text{attr}(i, j) = 
    \begin{cases} 
    1 & \text{if } j \text{ is the } k \text{ nearest neighbors of } i, \\
    0 & \text{otherwise.}
    \end{cases}
    \end{equation}
    \normalsize

    Next, the \abdisen\ performs message passing by using the adjacency matrix $\mathbf{A}_\text{attr}$ and node attributes $\mathbf{X}_\text{attr}$, generating the attribute bias embeddings (\textbf{\abemb}).
    This process inherently \textit{amplifies the attribute bias}, as the message-passing mechanism propagates and aggregates only the information related to node attributes in $\mathbf{X}_\text{attr}$ and $\mathbf{A}_\text{attr}$.

    The \abemb\ matrix $\mathbf{H}_\text{attr}^{(l+1)}\in \mathbb{R}^{n \times p}$ at layer $(l+1)$ is computed as follows:

    \small
    \begin{equation}
    \mathbf{H}_\text{attr}^{(l+1)} = \sigma\left(\mathbf{A}_\text{attr} \cdot \mathbf{H}_\text{attr}^{(l)} \cdot \mathbf{W}_\text{attr}^{(l)} + \mathbf{b}\right),
    \end{equation}
    \normalsize
    where $\mathbf{H}_\text{attr}^{(0)} = \mathbf{X}_\text{attr}$, $\mathbf{W}_\text{attr}^{(l)}$ indicates the learnable weight matrix at layer $l$. 
    $\sigma(\cdot)$ and $\mathbf{b} \in \mathbb{R}^{p}$ denote an
    activation function and a bias vector, respectively. 
    The final \abemb\ $\mathbf{H}_\text{attr}$ is obtained after $L$ layers of \abdisen\ GNN.

    \item \textbf{Structure Bias Disentangler (\sbdisen)}: 
    This component leverages another specialized GNN to capture the structure bias, where the message-passing mechanism updates node embeddings based solely on the graph's structure. 
    The process begins with the adjacency matrix $\mathbf{A}_\text{stru} \in \mathbb{R}^{n \times n}$, which represents the input graph's connectivity, and (randomly initialized) learnable node features $\mathbf{X}_\text{stru} \in \mathbb{R}^{n \times d}$. 
    By using these (random) features instead of actual node attributes, the \sbdisen\ ensures that \textit{only the structure bias is amplified}.
    
    The matrix of structure bias embeddings (\textbf{\sbemb})  $\mathbf{H}_\text{stru}^{(l+1)} \in \mathbb{R}^{n \times p}$ at layer $(l+1)$ is computed as follows:

    \small
    \begin{equation}
    \mathbf{H}_\text{stru}^{(l+1)} = \sigma\left(\mathbf{A}_\text{stru} \cdot \mathbf{H}_\text{stru}^{(l)} \cdot \mathbf{W}_\text{stru}^{(l)} + \mathbf{b}\right),
    \end{equation}
    \normalsize
    where $\mathbf{H}_\text{stru}^{(0)} = \mathbf{X}_\text{stru}$, and $\mathbf{W}_\text{stru}^{(l)}$ indicates the learnable weight matrix at layer $l$. The final \sbemb\ $\mathbf{H}_\text{stru}$ is obtained after $L$ layers of \sbdisen\ GNN.

    Note that the two GNNs used in the \abdisen\ and the \sbdisen\ have independent architectures and do not share their parameters (\eg, $\mathbf{W}_\text{attr}$ and $\mathbf{W}_\text{stru}$).
    This design allows the attribute and structure biases to be captured and amplified separately, minimizing any cross-contamination between these two distinct types of biases.

    \item \textbf{Potential Bias Disentangler (\pbdisen)}: 
    This component addresses the potential bias that arises from the interaction between attribute and structure biases.
    The process begins by creating an entangled embedding matrix $\mathbf{H}_\text{ent}$, which is generated by concatenating \abemb\ and \sbemb\ and passing them through a multi-layer perceptron (MLP):
    
    \small
    \begin{equation}
    \mathbf{H}_\text{ent} = \texttt{MLP}\left([\mathbf{H}_\text{attr} \,|\, \mathbf{H}_\text{stru}]\right),
    \end{equation}
    \normalsize
    where $[\cdot \,|\, \cdot]$ represents the concatenation of two matrices.

    The entangled embeddings preserve correlations between the two types of biases that might not be apparent individually.
    To produce the matrix of potential bias embeddings (\textbf{\pbemb}) $\mathbf{H}_\text{pot}\in \mathbb{R}^{n \times d}$, the \pbdisen\ subtracts \abemb\ and \sbemb\ from the entangled embeddings $\mathbf{H}_\text{ent}$:

    \small
    \begin{equation}
    \mathbf{H}_\text{pot} = \mathbf{H}_\text{ent} - \mathbf{H}_\text{attr} - \mathbf{H}_\text{stru}.
    \end{equation}
    \normalsize

    This subtraction effectively eliminates the individual impact of attribute and structure biases, revealing the more nuanced bias that emerges from their interaction.
\end{itemize}

After disentangling and amplifying each bias, the final step is to concatenate the disentangled embeddings into a comprehensive representation for each node, \ie, $\mathbf{H}_\text{final} = [\mathbf{H}_\text{attr} | \mathbf{H}_\text{stru} | \mathbf{H}_\text{pot}]$. 
This concatenated embedding
is then used to train the model for various downstream tasks.
For instance, in a node classification task, the loss function is typically the Negative Log-Likelihood (NLL) loss~\cite{bishop2007}: 

\small
\begin{equation}
\mathcal{L}_\text{primary} = - \sum_{i=1}^{n} [\mathbf{y}_{i}\log(\hat{\mathbf{y}}_i) + (1-\mathbf{y}_{i})\log(1-\hat{\mathbf{y}}_i)],
\end{equation}
\normalsize
where $\mathbf{y}_i$ and $\hat{\mathbf{y}}_i$ indicate the true label of node $i$ and the predicted probability for the label of node $i$, respectively. 
For other tasks, the appropriate loss function should be applied.

\vspace{1mm}
\noindent\textbf{(M2) Debiasing Module}.
The goal of this module is to refine the disentangled embeddings to ensure that predictions are free from biases related to sensitive attributes.
Despite the initial disentanglement, the embeddings for different bias types may still overlap in the embedding space due to residual similarities or interdependencies.
Thus, it is crucial to achieve a clear separation of each bias in the embedding space and most importantly, to eliminate any sensitive information from the corresponding embeddings. 
To do this, this module leverages two regularizers: the bias contrast optimizer (BCO) and the fairness harmonizer (FH).

\begin{itemize}
\item \textbf{Bias Contrast Optimizer (BCO)}: 
This component ensures that the three types of disentangled embeddings--attribute, structure, and potential biases--remain distinct and clearly represent their respective bias.
The regularizer $\mathcal{L}_\text{bco}$ for this process is formally defined as:

\vspace{-0.2cm}
\small
\begin{equation}
\mathcal{L}_\text{bco} = - \sum_{q \neq r} \texttt{D}_{f}\left(\mathbf{H}_q, \mathbf{H}_r\right),
\end{equation}
\normalsize
where $q, r \in \{\text{attr}, \text{stru}, \text{pot}\}$. 
The distance function $\texttt{D}_{f}(\cdot, \cdot)$ is defined by using the Frobenius norm~\cite{golub2013matrix} as follows:

\small
\begin{equation}
\texttt{D}_{f}\left(\mathbf{H}_q, \mathbf{H}_r\right) = |\mathbf{H}_q - \mathbf{H}_r|_F.
\end{equation}
\normalsize
This regularizer enforces a strong separation between embeddings from different bias components.

\item \textbf{Fairness Harmonizer (FH)}: 
This component reduces sensitive information in the disentangled embeddings by minimizing the Wasserstein-1 distance~\cite{villani2003topics} between subgroup distributions for each bias type $q$.
The regularizer $\mathcal{L}_\text{fh}$ is formally defined as follows:

\small
\begin{equation}
\mathcal{L}_\text{fh} = \sum_{q\in \{\text{attr}, \text{stru}, \text{pot}\}} \texttt{W}\left(\mathcal{P}(\mathbf{H}_q(0)), \mathcal{P}(\mathbf{H}_q(1))\right),
\end{equation}
\normalsize
where $\texttt{W}(\cdot, \cdot)$ denotes the Wasserstein distance between two probability distributions $\mathcal{P}(\mathbf{H}_q(0))$ and $\mathcal{P}(\mathbf{H}_q(1))$, representing the distributions of disentangled embeddings for the bias type $q$ in subgroups 0 and 1, respectively. 
The Wasserstein distance $\texttt{W}(\mathcal{P}, \mathcal{Q})$ is calculated as follows:

\small
\begin{equation}
\texttt{W}(\mathcal{P}, \mathcal{Q}) = \inf_{\gamma \in \Gamma(\mathcal{P}, \mathcal{Q})} \mathbb{E}_{(x,y) \sim \gamma} [\|x - y\|_1],
\end{equation}
\normalsize
where $\Gamma(\mathcal{P}, \mathcal{Q})$ denotes the set of all joint distributions $\gamma(x,y)$ whose marginals are $\mathcal{P}$ and $\mathcal{Q}$.
However, since directly calculating the Wasserstein distance is intractable, we adopted an approximation from ~\citet{DongLJL22} to enable end-to-end gradient optimization.

\end{itemize}

It should be noted that $\mathcal{L}_\text{bco}$ operates across different types of bias embeddings--\abemb, \sbemb, and \pbemb, while $\mathcal{L}_\text{fh}$ refines the embeddings within each bias type.

\subsection{Training}

For each node $i$ in the input graph $\mathcal{G}$, its disentangled embeddings and associated parameters (\eg, $\mathbf{W}_\text{attr}$ and $\mathbf{W}_\text{stru}$) are learned by optimizing the following loss function:

\small
\begin{equation}
\mathcal{L}_{\text{total}} = \mathcal{L}_\text{primary} + \alpha \cdot \mathcal{L}_\text{fh} + \beta \cdot \mathcal{L}_\text{bco},
\end{equation}
\normalsize
where $\alpha$ and $\beta$ denote hyperparameters that balance the contributions of the FH and BCO regularizers, respectively.
By employing this training approach, \ours\ ensures that the learned embeddings not only achieve high accuracy for the primary downstream task but also maintain fairness, leading to more equitable and reliable predictions.

\begin{table*}[t]
\centering
\renewcommand{\arraystretch}{1.15}
\resizebox{\textwidth}{!}{%
\begin{tabular}{cc|cc|cccccccccc|c}
\toprule
& \textbf{Metrics} & \textbf{L1-Vanilla} & \textbf{L3-Vanilla} & \textbf{FairGNN} & \textbf{NIFTY} & \textbf{EDITS} & \textbf{FairVGNN} & \textbf{CAF} & \textbf{GEAR} & \textbf{BIND} & \textbf{PFR-AX} & \textbf{PostProcess} & \textbf{FairSIN} & \textbf{\ours}  \\
\midrule
\multirow{6}{*}{\rotatebox{90}{\textbf{NBA}}} &
\textbf{ACC ($\uparrow$)} & 57.97 & 58.73 & 60.76 & 63.29 & 69.11  & 65.57  & 60.51  & 57.98  & 60.76  & \underline{70.63} & 58.73  & 66.58 & \textbf{71.39} \\
& \textbf{AUC ($\uparrow$)} & 63.75  & 63.33  & 74.91  & 70.75  & 71.82  & \underline{79.96} & 67.06  & 60.04  & 79.33  & 73.26  & 63.33  & 71.72  & \textbf{80.56} \\
& \textbf{F1 ($\uparrow$)} & 61.55  & 62.00 & 70.69  & 66.86  & \textbf{74.99} & 72.93  & 68.81  & 65.08  & 70.50  & 74.06  & 62.00  & \underline{74.21} & 73.51  \\ \cmidrule{2-15}
& \textbf{SP ($\downarrow$)} & 32.94  & 32.83  & 6.39  & 9.82 & 8.98  & 7.82 & \textbf{0.00} & 20.53  & 4.55  & 4.03  & 32.83  & 12.96  & \underline{1.12 } \\
& \textbf{EO ($\downarrow$)} & 33.68  & 35.95  & 10.14  & 8.60  & 4.39  & 13.28  & \textbf{0.00} & 21.94 & 1.77 & 13.56  & 35.95  & 2.34  & \underline{0.80 } \\
\midrule
\multirow{6}{*}{\rotatebox{90}{\textbf{Recidivism}}} &
\textbf{ACC ($\uparrow$)} & 84.18  & 83.73  & 84.50  & 79.94  & 78.18  & 83.64  & \underline{86.79} & 78.32  & 84.49  & 85.41  & 81.28  & 86.59  & \textbf{89.99 } \\
& \textbf{AUC ($\uparrow$)} & 86.90  & 86.84  & 89.05  & 81.23  & 83.62  & 84.38  & 87.07  & 81.30  & 89.13  & \underline{89.48 } & 83.23  & 89.08  & \textbf{93.41 } \\
& \textbf{F1 ($\uparrow$)} & 78.65  & 78.10  & 79.77  & 69.77  & 73.16  & 76.89  & 80.63  & 71.18  & 79.82  & 79.48  & 75.91  & \underline{80.87} & \textbf{86.31 } \\ \cmidrule{2-15}
& \textbf{SP ($\downarrow$)} & 7.79  & 8.13 & 6.64  & 3.69 & 10.89 & 5.42 & 5.73  & 5.81  & 9.24  & 6.13  & \underline{1.43} & 5.65  & \textbf{0.73} \\
& \textbf{EO ($\downarrow$)} & 5.23  & 5.65  & 3.16  & 2.97  & 7.62  & 3.92  & 3.41  & 4.11  & 4.61  & 4.14  & \underline{2.92} & 3.59  & \textbf{0.90} \\
\midrule
\multirow{6}{*}{\rotatebox{90}{\textbf{Credit}}} &
\textbf{ACC ($\uparrow$)} & 73.57  & 73.92  & 73.99  & 73.43  & 74.77  & \underline{77.92} & 76.00  & o.o.m & 74.60  & 63.96 & 73.21  & 77.60 & \textbf{78.19} \\
& \textbf{AUC ($\uparrow$)} & \textbf{73.48} & \underline{73.40} & 64.19  & 72.14  & 72.30  & 68.67  & 65.72  & o.o.m & 71.91  & 66.90 & 70.10  & 71.57  & 71.41  \\
& \textbf{F1 ($\uparrow$)} & 81.87  & 82.16  & 83.08  & 81.70  & 82.99  & \textbf{87.48} & 85.15  & o.o.m & 82.76  & 73.95 & 82.03  & 87.23  & \underline{87.39} \\ \cmidrule{2-15}
& \textbf{SP ($\downarrow$)} & 13.88 & 12.18  & 3.17  & 11.60  & 7.98  & \textbf{0.40} & 11.70  & o.o.m & 11.76  & 19.19  & 1.39  & 0.69  & \underline{0.44} \\
& \textbf{EO ($\downarrow$)} & 11.68  & 10.04 & 1.73  & 9.30  & 6.09  & \textbf{0.16} & 8.51  & o.o.m & 9.15  & 22.66  & 1.83  & 0.66  & \underline{0.45} \\
\midrule
\multirow{6}{*}{\rotatebox{90}{\textbf{Pokec\_n}}} &
\textbf{ACC ($\uparrow$)} & 66.97  & 65.27  & 63.56  & \underline{67.86} & o.o.m & \textbf{69.51} & o.o.m & o.o.m & 55.69  & o.o.m & 66.54  & 65.69  & 67.18  \\
& \textbf{AUC ($\uparrow$)} & 72.73 & 70.74  & 67.10  & \underline{73.92} & o.o.m & \textbf{73.99} & o.o.m & o.o.m & 58.99  & o.o.m & 71.76  & 72.89  & 73.68  \\
& \textbf{F1 ($\uparrow$)} & 65.70 & 64.91  & 59.79 & \underline{66.25} & o.o.m & 66.01 & o.o.m & o.o.m & 52.36 & o.o.m & 65.91 & \textbf{67.44} & 62.34 \\ \cmidrule{2-15}
& \textbf{SP ($\downarrow$)} & 7.90 & 17.19 & 3.28  & \underline{1.20} & o.o.m & 2.77  & o.o.m & o.o.m & 6.78 & o.o.m & 14.97 & 2.40  & \textbf{0.71} \\
& \textbf{EO ($\downarrow$)} & 7.09 & 14.88 & 5.05 & \underline{1.23} & o.o.m & 3.38 & o.o.m & o.o.m & 5.96 & o.o.m & 11.38  & 1.64 & \textbf{1.09} \\
\midrule
\multirow{6}{*}{\rotatebox{90}{\textbf{Pokec\_z}}} &
\textbf{ACC ($\uparrow$)} & 64.92  & 65.40 & 62.97  & \underline{65.71} & o.o.m & 63.38 & o.o.m & o.o.m & 58.38 & o.o.m & 64.39  & 62.21  & \textbf{68.56} \\
& \textbf{AUC ($\uparrow$)} & 70.03  & 69.84 & 65.81  & \underline{70.57} & o.o.m & 68.99 & o.o.m & o.o.m & 61.20 & o.o.m & 69.08  & 68.81  & \textbf{74.85} \\
& \textbf{F1 ($\uparrow$)} & 65.48 & 65.08  & 64.47  & 65.00  & o.o.m & \underline{67.31} & o.o.m & o.o.m & 58.13 & o.o.m & 65.45  & 65.37  & \textbf{67.94} \\ \cmidrule{2-15}
& \textbf{SP ($\downarrow$)} & 7.27 & 10.91  & 4.79  & 5.03  & o.o.m & 5.04 & o.o.m & o.o.m & 6.13 & o.o.m & 12.18  & \underline{0.96} & \textbf{0.67} \\
& \textbf{EO ($\downarrow$)} & 4.05 & 7.88  & 3.65  & \underline{1.24} & o.o.m & 3.06 & o.o.m & o.o.m & 4.96 & o.o.m & 7.14 & 1.64  & \textbf{0.73} \\
\bottomrule
\end{tabular}%
}
\caption{Accuracy and fairness results of \ours\ and competitors across five real-world datasets. ($\uparrow$) and ($\downarrow$) mean higher and lower values are better, respectively; `o.o.m' denotes `out of memory.'} 
\label{tab:comparison}
\end{table*}

\section{Evaluation}\label{sec:evaulation}

We designed our experiments, aiming at answering the following key evaluation questions (EQs):

\begin{itemize}[leftmargin=*]
    \item \textbf{(EQ1)} Does \ours\ outperform competitors in balancing accuracy and fairness?
    \item \textbf{(EQ2)} What is the impact of bias disentangling, amplifying, and debiasing strategies on model performance?
    \item \textbf{(EQ3)} How well are the different biases isolated in the embedding space?
    \item \textbf{(EQ4)} How sensitive is the performance of \ours\ to the hyperparameters $\alpha$ and $\beta$? 
\end{itemize}

We also showed that \ours\ requires a reasonable computational cost, with training time increasing approximately linearly as the number of nodes grows. Detailed experimental results are available in the online appendix at \url{https://github.com/Bigdasgit/DAB-GNN}.

\subsection{Experimental Setup}

\vspace{1mm}
\noindent\textbf{Datasets}.
We used 5 real-world graph datasets for our experiments: NBA, Recidivism, Credit, Pokec\_n, and Pokec\_z, which are all publicly available.
Table~\ref{tab:dataset} provides key statistics for these datasets (more details in Section~\ref{sec:pre}).

\vspace{1mm}
\noindent\textbf{Competitors}.
We compared \ours\ against two baselines--Vanilla GCN with one layer (L1-Vanilla) and three layers (L3-Vanilla)--as well as ten state-of-the-art \fagnn\ methods: FairGNN~\cite{DaiW21}, NIFTY~\cite{AgarwalLZ21}, EDITS~\cite{DongLJL22}, FairVGNN~\cite{WangZDCLD22}, CAF~\cite{Guo0X0W23}, GEAR~\cite{MaGWYZL22}, BIND~\cite{DongW0LL23}, PFR-AX~\cite{MerchantC23}, PostProcess~\cite{MerchantC23}, and FairSIN \cite{YangLYS24}.
We used the source codes provided by the authors.

\vspace{1mm}
\noindent\textbf{Evaluation Tasks}.
Following previous studies~\cite{AgarwalLZ21, DaiW21, DongLJL22, WangZDCLD22}, we assessed the methods by using a node classification task, splitting the nodes into training (50\%), validation (25\%), and test (25\%) sets. 
We measured accuracy with three metrics: Accuracy (ACC), Area Under the Curve (AUC), and F1-Score.
In addition, we evaluate fairness by using \textit{Statistical Parity} (SP)~\cite{DworkHPRZ12} and \textit{Equality of Opportunity} (EO)~\cite{HardtPNS16}, where lower values indicate better model fairness.
Detailed equations for calculating SP and EO are provided in the online appendix.

\vspace{1mm}
\noindent\textbf{Implementation Details}.
We used Vanilla GCN as the backbone for both the competitors and \ours, carefully tuning their hyperparameters via grid search.
For \ours, we used a 3-layer GCN with a hidden layer of size 16, a gradient penalty~\cite{GulrajaniAADC17} of 10, 
1,000 epochs, the embedding dimensionality of 48, and a weight decay of 0.00001.
The hyperparameters $k$ for $k$-NN graph construction and $\alpha$ for $\mathcal{L}_\text{fh}$ were tuned within the range of \{1, 100\}, while $\beta$ for $\mathcal{L}_\text{bco}$ was tuned within \{0.00001, 0.1\}. 
All experiments were conducted by using five different seed settings, and we report the average accuracy.
For complete implementation details, please refer to the online appendix.

\subsection{Results}

\noindent\textbf{(EQ 1) Comparison with 12 Competitors}.
To evaluate how well \ours\ balances accuracy and fairness, we compared it against two baselines and ten state-of-the-art \fagnn\ methods. 
In Table~\ref{tab:comparison}, \textbf{boldface} and \ul{underlined} values indicate the best and 2nd-best performance in each row, respectively. 
Higher ACC, AUC, and F1-score values indicate better accuracy, while lower SP and EO values indicate better fairness. 

\ours\ shows substantial improvements over the \textit{baselines} in almost all cases.
For example, on the Recidivism dataset, \ours\ improves AUC by about 7.57\% and reduces SP by around 91.02\% compared to L3-Vanilla. 
This demonstrates that \ours\ not only addresses fairness effectively but also enhances accuracy, which is typically challenging under fairness constraints.
Compared to \textit{state-of-the-art \fagnn\ methods}, \ours\ consistently achieves the best balance between accuracy and fairness. 
Although there are a few cases where \ours\ may have slightly lower accuracy or higher fairness values than certain competitors, these competitors often sacrifice significantly one metric for the other. 
For instance, CAF on the NBA dataset achieves perfect fairness metrics (SP and EO of 0), but this comes at the cost of a significant drop in accuracy. 
This trade-off underscores the difficulty in balancing accuracy and fairness.

\begin{table}[t]
\centering
\renewcommand{\arraystretch}{1.15}
\resizebox{0.49\textwidth}{!}{%
\begin{tabular}{cc|ccc|cc|c}
\toprule
& \multirow{2}{*}{\textbf{Metrics}} & \multicolumn{3}{c|}{\textbf{(a) Disen. and Amp. Module}} & \multicolumn{2}{c|}{\textbf{(b) Debiasing Module}} & \multirow{2}{*}{\textbf{\ours}} \\
& & \textbf{w/o \abdisen} & \textbf{w/o \sbdisen} & \textbf{w/o \pbdisen} & \textbf{w/o $\mathcal{L}_\text{fh}$} & \textbf{w/o $\mathcal{L}_\text{bco}$} &  \\
\midrule
\multirow{6}{*}{\rotatebox{90}{\textbf{NBA}}} & \textbf{ACC ($\uparrow$)}  & 70.38 & 70.89  & \textbf{71.65} & 69.11  & 69.37 & \underline{71.39} \\
 & \textbf{AUC ($\uparrow$)}  & 79.51  & 81.13  & \textbf{81.86} & 77.68 & \underline{81.14} & 80.56  \\
 & \textbf{F1 ($\uparrow$)}  & 73.14  & \textbf{74.07} & 73.36  & 70.48  & 72.05 & \underline{73.51} \\ \cmidrule{2-8}
 & \textbf{SP ($\downarrow$)}  & 6.08  & \underline{5.97} & 6.52  & 9.50  & 6.76  & \textbf{1.12} \\
 & \textbf{EO ($\downarrow$)}  & 8.66 & 8.26  & 12.25  & 12.65  & \underline{8.03} & \textbf{0.80} \\
\midrule
\multirow{6}{*}{\rotatebox{90}{\textbf{Recidivism}}} & \textbf{ACC ($\uparrow$)}  & 88.82 & \underline{91.59} & 91.25  & \textbf{91.63} & 91.01  & 89.99  \\
 & \textbf{AUC ($\uparrow$)}  & 92.38 & \underline{94.13} & 94.07  & \textbf{94.24} & 94.04  & 93.41 \\
 & \textbf{F1 ($\uparrow$)}  & 84.20  & \textbf{88.17} & 87.72  & \underline{87.73} & 87.53 & 86.31 \\ \cmidrule{2-8}
 & \textbf{SP ($\downarrow$)}  & 1.13 & 1.21  & 1.03  & 3.30  & \underline{0.86} & \textbf{0.73} \\
 & \textbf{EO ($\downarrow$)}  & \textbf{0.77} & 2.18  & 1.45  & 2.26  & 1.61  & \underline{0.90} \\
\midrule
\multirow{6}{*}{\rotatebox{90}{\textbf{Credit}}} & \textbf{ACC ($\uparrow$)}  & 74.22 & 75.06  & 75.75  & \underline{76.47} & 74.10  & \textbf{78.19} \\
 & \textbf{AUC ($\uparrow$)}  & \textbf{72.86} & 68.45  & \underline{72.05} & 64.08  & 69.66  & 71.41  \\
 & \textbf{F1 ($\uparrow$)}  & 82.84  & 84.04  & 84.80  & \underline{85.93} & 83.16 & \textbf{87.39} \\ \cmidrule{2-8}
 & \textbf{SP ($\downarrow$)} & 6.89  & 5.68  & 4.73    & \underline{2.59} & 5.98  & \textbf{0.44} \\
 & \textbf{EO ($\downarrow$)}  & 5.67  & 4.90  & 3.76  & \underline{2.30} & 5.12  & \textbf{0.45} \\
\midrule
\multirow{6}{*}{\rotatebox{90}{\textbf{Pokec\_n}}} & \textbf{ACC ($\uparrow$)}  & 61.18  & \underline{67.33} & 66.50  & 66.79  & \textbf{67.75} & 67.18  \\
 & \textbf{AUC ($\uparrow$)}  & 67.87 & 72.89 & 72.96 & \underline{73.62} & 72.82  & \textbf{73.68} \\
 & \textbf{F1 ($\uparrow$)}  & 58.22 & 61.48  & \underline{63.85} & \textbf{64.54} & 62.78  & 62.34 \\ \cmidrule{2-8}
 & \textbf{SP ($\downarrow$)}  & 2.03  & 2.21  & 2.04  & 4.05  & \underline{1.38} & \textbf{0.71} \\
 & \textbf{EO ($\downarrow$)}  & \underline{2.57} & 3.30  & 2.64  & 5.40  & 3.61  & \textbf{1.09} \\
\midrule
\multirow{6}{*}{\rotatebox{90}{\textbf{Pokec\_z}}} & \textbf{ACC ($\uparrow$)}  & 67.32 & \underline{68.92} & 68.82  & \textbf{69.02} & 67.99  & 68.56  \\
 & \textbf{AUC ($\uparrow$)}  & 73.70  & \underline{75.14} & 74.49 & \textbf{75.39} & 75.05  & 74.85  \\
 & \textbf{F1 ($\uparrow$)}  & 65.89 & 67.60  & 67.16 & \underline{68.30} & \textbf{69.17} & 67.94  \\ \cmidrule{2-8}
 & \textbf{SP ($\downarrow$)}  & 3.55  & \underline{1.88} & 4.82  & 4.59  & 2.43  & \textbf{0.67} \\
 & \textbf{EO ($\downarrow$)}  & 3.70  & \underline{1.94} & 3.97  & 3.61  & 2.51 & \textbf{0.73} \\
\bottomrule
\end{tabular}%
}
\caption{The ablation studies for \ours.}
\label{tab:ablation}
\end{table}

\vspace{1mm}
\noindent\textbf{(EQ 2) Ablation Studies}.
We performed ablation studies to evaluate the impact of excluding specific components from the key modules in \ours.
The variants include: (a) Disentanglement and Amplification--w/o \abdisen, w/o \sbdisen, and w/o \pbdisen; and (b) Debiasing--w/o $\mathcal{L}_\text{fh}$ and w/o $\mathcal{L}_\text{bco}$. 

In terms of \textbf{accuracy}, removing a specific component from the key modules can slightly improve accuracy in some cases. 
However, \ours\ generally maintains accuracy comparable to the best-performing variants. 
Additionally, the most impactful disentangler or regularizer for accuracy varies across datasets, highlighting \ours's ability to achieve generalized accuracy across diverse datasets by addressing all biases comprehensively.
In terms of \textbf{fairness}, \ours\ achieves the best results across all variants (except for 2nd best in EO on Recidivism), underscoring the crucial role of disentangling, amplifying, and debiasing biases to achieve fair outcomes. 
This analysis highlights the importance of each module in \ours\ for balancing accuracy and fairness.

\begin{figure}[t]
    \centering
    \begin{subfigure}[b]{0.22\textwidth}
        \centering
        \includegraphics[width=\textwidth]{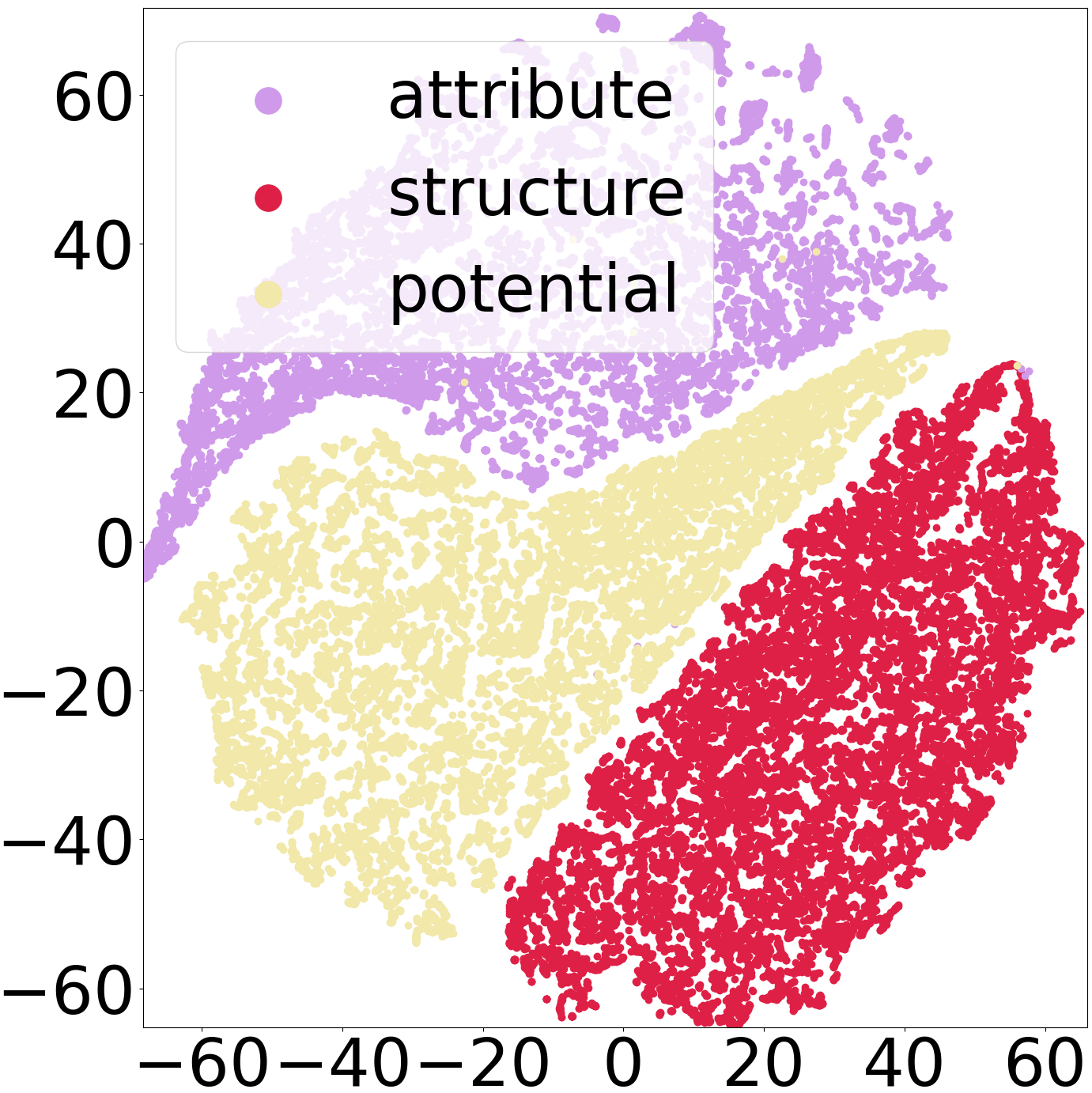}
        \caption{\textbf{Recidivism}}
        \label{fig:bail_tsne}
    \end{subfigure}
    \begin{subfigure}[b]{0.22\textwidth}
        \centering
        \includegraphics[width=\textwidth]{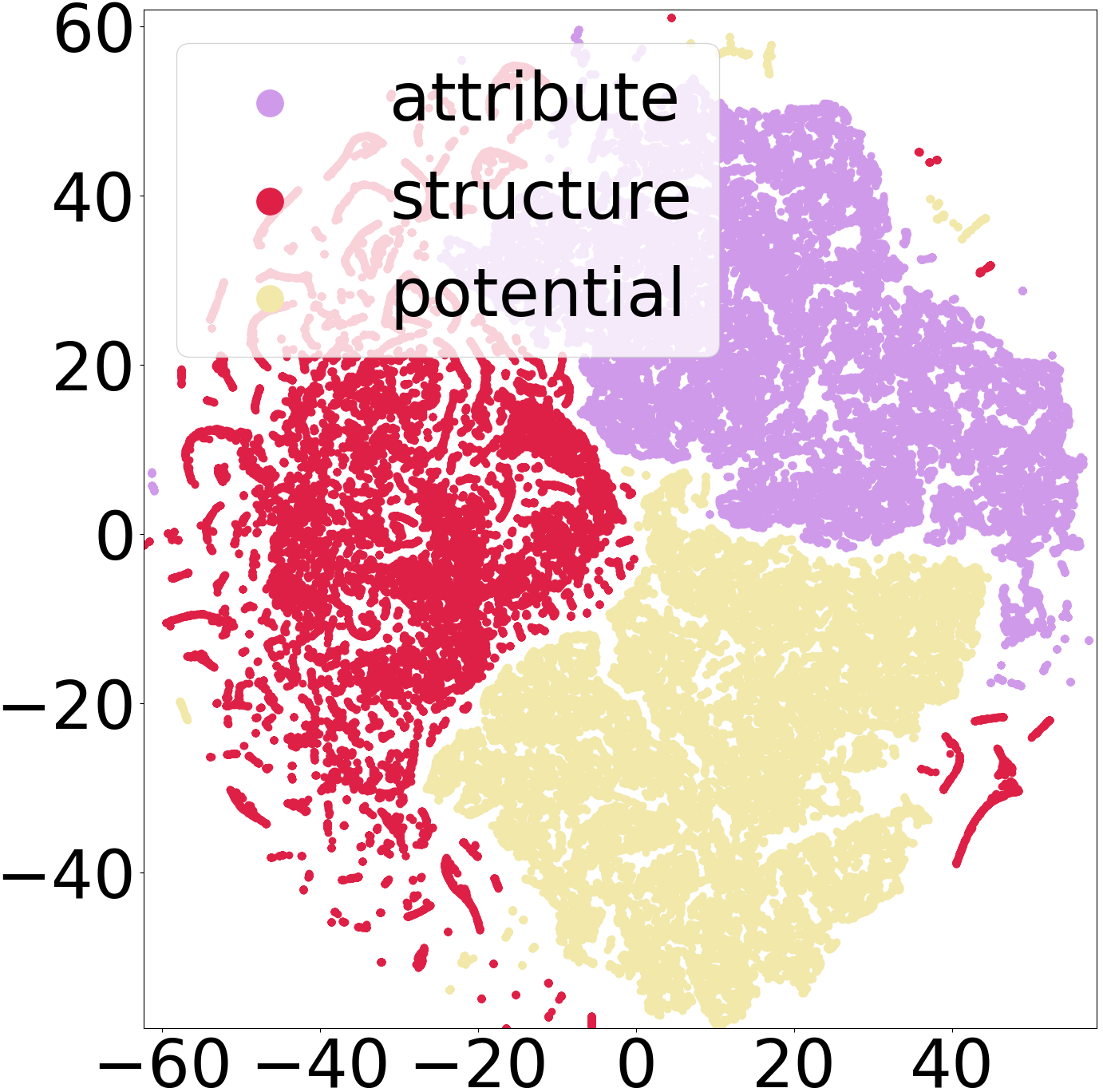}
        \caption{\textbf{Credit}}
        \label{fig:credit_tsne}
    \end{subfigure}
    
    \caption{Visualization of disentangled node embeddings by using t-SNE: \abemb, \sbemb, and \pbemb.}
    \label{fig:tsne_grid}
\end{figure}

\vspace{1mm}
\noindent\textbf{(EQ 3) Disentangled Embeddings Analysis}.
As shown in Figure~\ref{fig:tsne_grid}, we visualized the final disentangled embeddings using t-SNE~\cite{vandermaaten08a}.
Different colors indicate different types of bias embeddings: purple for \abemb, red for \sbemb, and yellow for \pbemb.
The visualizations clearly show that the attribute, structure, and potential bias embeddings are well-separated into distinct clusters, demonstrating that our disentanglement process successfully isolates the various biases present in the graph data. 

\begin{figure}[t]
\footnotesize
    \centering
    \begin{tabular}{cc}
     \includegraphics[width=0.45\linewidth,height=2.2cm]{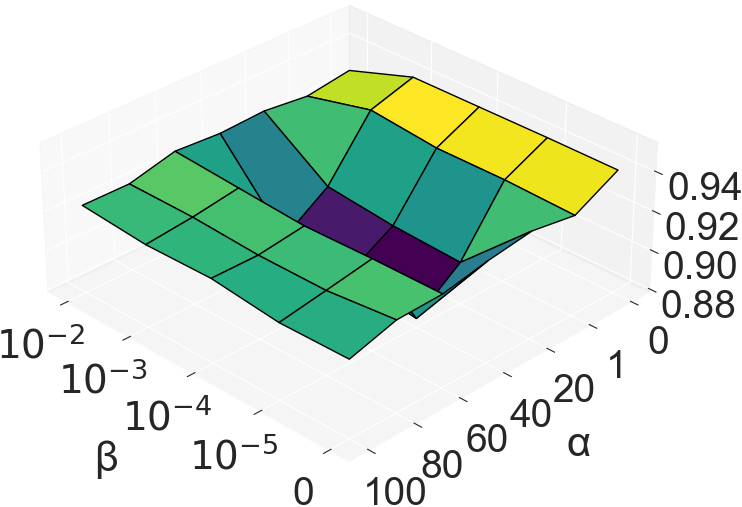} &         \includegraphics[width=0.45\linewidth,height=2.2cm]{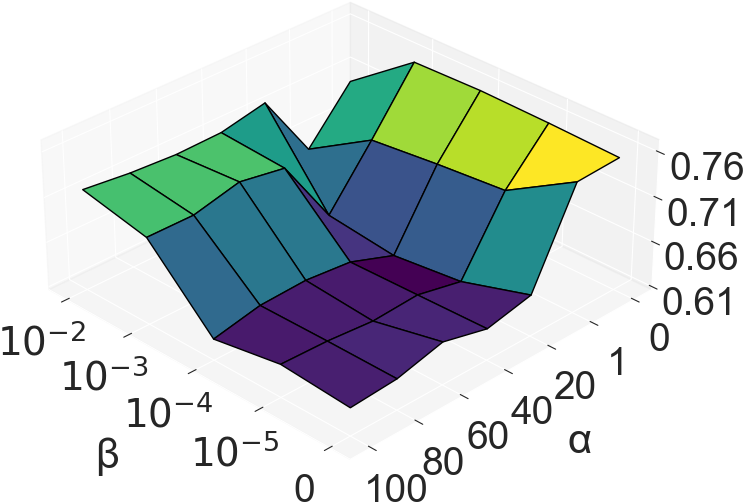} \\
    \multicolumn{2}{c}{(i) AUC}\\
    \includegraphics[width=0.45\linewidth,height=2.2cm]{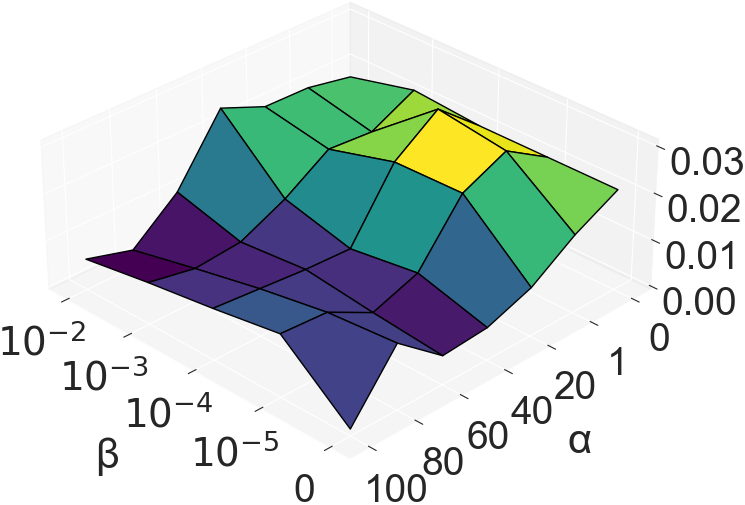} & 
    \includegraphics[width=0.45\linewidth,height=2.2cm]{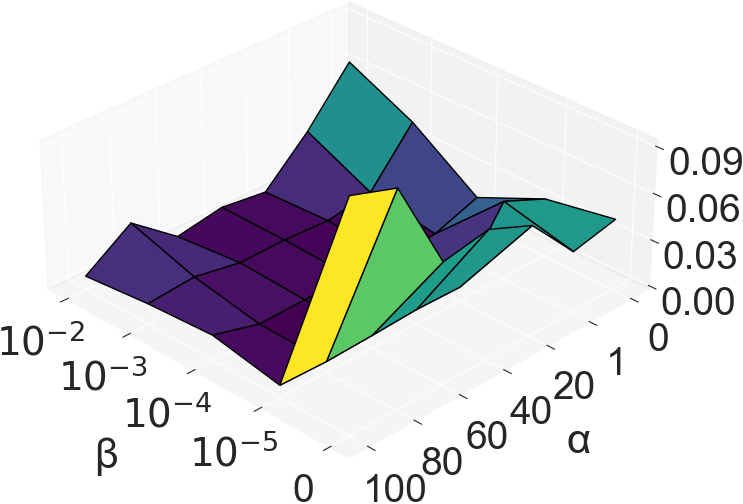} \\
    \multicolumn{2}{c}{(ii) EO} \\
    (a) \textbf{Recidivism} & (b) \textbf{Credit} \\   
    \end{tabular}
    \caption{The effects of $\alpha$ and $\beta$ on AUC ($\uparrow$) and EO ($\downarrow$).}
    \label{fig:alpha_beta_auc_sp}
\end{figure}

\vspace{1mm}
\noindent\textbf{(EQ 4) Hyperparameters Analysis}.
To evaluate the sensitivity of the performance with \ours\ to the hyperparameters $\alpha$ (for $\mathcal{L}_\text{fh}$) and $\beta$ (for $\mathcal{L}_\text{bco}$), we analyzed their impact on AUC for accuracy and EO for fairness on the Recidivism and Credit datasets (see Figure~\ref{fig:alpha_beta_auc_sp}). 
In both datasets, high values for $\alpha$ and $\beta$ generally provide the best balance between accuracy and fairness, highlighting the importance of both regularizers.
The optimal ranges are approximately greater than 60 for $\alpha$ and 0.001 for $\beta$.

\section{Conclusion}
In this work, we identified a significant challenge inherent to existing fairness-aware GNN methods: the entanglement of different bias types in the final node embeddings leads to difficulty in their comprehensive debiasing. 
To address this challenge, we introduced \ours, a novel GNN framework that disentangles, amplifies, and debiases the attribute, structure, and potential biases within node embeddings. 
Extensive experiments on five real-world graph datasets show that \ours\ outperforms ten state-of-the-art competitors in balancing accuracy and fairness, while validating the effectiveness of our design choices. 

\section*{Acknowledgments}
The work of Sang-Wook Kim was supported by the Institute of Information \& communications Technology Planning \& Evaluation (IITP) grant funded by the Korea government(MSIT) (No.2022-0-00352, No.RS-2022-00155586; A High-Performance Big-Hypergraph Mining Platform for Real-World Downstream Tasks).
Yeon-Chang Lee's work was supported by the Institute of Information \& communications Technology Planning \& Evaluation (IITP) grant, funded by the Korea government (MSIT) (No. RS-2020-II201336, Artificial Intelligence Graduate School Program (UNIST)).

\bibliography{aaai25}

\end{document}